\pdfoutput=1

\documentclass[11pt]{article}

\usepackage[]{acl}

\usepackage{times}
\usepackage{latexsym}
\usepackage{graphicx}
\usepackage{float}
\usepackage{graphics}

\usepackage[T1]{fontenc}

\usepackage[utf8]{inputenc}

\usepackage{microtype}

\title{Meeting Summarization: A Survey of the State of the Art}

\author{Lakshmi Prasanna Kumar \\
  IMRSV Data Labs., Ottawa, Canada \\
  \texttt{lakshmi@imrsv.ai} \\\And
  Arman Kabiri \\
  IMRSV Data Labs., Ottawa, Canada \\
  \texttt{arman@imrsv.ai} \\}

\begin{document}
\maketitle
\begin{abstract}
Information overloading requires the need for summarizers to extract salient information from the text. Currently, there is an overload of dialogue data due to the rise of virtual communication platforms. The rise of Covid-19 has led people to rely on online communication platforms like Zoom, Slack, Microsoft Teams, Discord, etc. to conduct their company meetings. Instead of going through the entire meeting transcripts, people can use meeting summarizers to select useful data. Nevertheless, there is a lack of comprehensive surveys in the field of meeting summarizers. In this survey, we aim to cover recent meeting summarization techniques. Our survey offers a general overview of text summarization along with datasets and evaluation metrics for meeting summarization. We also provide the performance of each summarizer on a leaderboard. We conclude our survey with different challenges in this domain and potential research opportunities for future researchers.
\end{abstract}

\section{Introduction}

In the current pandemic era, numerous meetings are happening virtually thereby producing substantial amount of meeting data on a daily basis. Meeting participants need to do the laborious task of going through the meeting transcripts to extract the salient information which in turn has attracted researchers attention to the need of effective meeting summarizers. According to \citeauthor{feng2021survey}, meeting summarization, the task of summarizing online multi-participant meetings, is a difficult task to perform due to its multi-speaker structure and dialogue style which lead to low information density. The fact that multiple participants in a meeting have different language styles and roles result in meeting heterogeneity. There are other challenges like lack of annotated datasets, length of the meeting transcripts, etc. This paper performs a thorough study of over forty papers covering different summarization techniques for meetings and discuss the challenges in this field.

The paper starts by giving a background on the standard text summarization techniques followed by dialogue summarization and the metrics used for evaluation in section \ref{sec:background}. In section \ref{sec:meetingsummarizers}, we discuss the meeting summarization techniques followed by the datasets commonly used for training and evaluation of the meeting summarization models in section \ref{sec:datasets}. The discussions and challenges faced in meeting summarization is explained in sections \ref{sec:Discussions} and \ref{sec:challenges}, respectively. We conclude the survey in section \ref{sec:conclusion} by giving future research directions for the potential researchers who are interested to work in this domain.

\section{Background}
\label{sec:background}
In this section we briefly describe the automatic text summarization methods followed by an overview of dialogue summarization tasks. A brief discussion of evaluation metrics is also discussed here. 

\subsection{Automatic Text Summarization}
Automatic text summarization is an inevitable task to process and understand vast amount of textual data which is available in the form of web contents, scientific papers, legal documents, news articles, medical documents, etc. According to~\cite{hovy2005automated}, a summary is defined as a text that is produced out of one or more texts that contain a significant portion of the information of the original text(s), and that is no longer than half of the original text(s). It is a difficult task for computers to understand the entire context of a document and find the significant data in it as compared to humans~\cite{allahyari2017text}. Scientists have started the research in text summarization as early as 1950, and they are still seeking new techniques to achieve a summary as close to human summary~\cite{gambhir2017recent}.

Automatic text summarization(ATS) can be classified in different ways~\cite{el2021automatic}. Some of the important categories are discussed below. ATS can be single document summarization or multi-document summarization depending on the type of input source. A multi-document summarization system extracts salient information from multiple documents according to some topic and presents them in a coherent concise manner~\cite{lin2002single}. 
Besides, based on the summarization approach, ATS can be divided into extractive and abstractive summarization. Extractive approach extracts the most important sentences from the input text and generates the summary by concatenating them, whereas abstractive approach generates a summary by paraphrasing the contents of the input text~\cite{nallapati2017summarunner}. Most of the time, extractive summaries are simpler and more accurate than the latter. Maximal Marginal Relevance (MMR) algorithm suggested by \citeauthor{carbonell1998use} was one of the initial methods for sentence selection in extractive text summarization followed by integer linear programming (ILP)~\cite{mcdonald2007study}, submodular-based approaches~\cite{lin2012learning}, etc. For sentence scoring, there are graph-based approaches like LexRank~\cite{erkan2004lexrank} and TextRank~\cite{mihalcea2004textrank}. Some neural network-based approaches are also proposed for sentence scoring~\cite{cao2015ranking,ren2017leveraging}. The following years saw a rise in popularity of using neural network-based models for both extractive and abstractive paradigms. \citeauthor{nallapati2016abstractive} and \citeauthor{nallapati2017summarunner} proposed abstractive and extractive techniques using recurrent neural networks(RNN) for generating multi-sentence summaries. 
In order to overcome the repetition of words and to increase the accuracy, \citeauthor{see2017get} suggested a hybrid pointer-generator network with coverage mechanism. With the introduction of transformer-based sequence-to-sequence models~\cite{vaswani2017attention}, there is a boom in the use of pretrained language models for text summarization tasks. The transformer architecture enables transfer learning for NLP by training language models on an unlabeled raw corpus and then fine-tuning for specific downstream tasks. Pretrained language models like Bert~\cite{liu2019text}, GPT~\cite{radford2019language}, Bart~\cite{lewis2019bart}, and Pegasus~\cite{zhang2020pegasus} have been very successful in this domain.

\subsection{Dialogue Summarization}
Most of the published work on text summarization has targeted single-speaker formal documents like news, scientific articles, etc~\cite{cachola2020tldr,yasunaga2019scisummnet,narayan2018don,zhang2020pegasus}. Dialogue summarization is a domain which is gaining popularity due to the availability of high volume conversational data obtained when people use digital platforms and smartphones to exchange information. Dialogue domain can contain emails, meetings, online chats, customer service interactions, medical conversations between doctors and patients, podcasts, etc.~\cite{feng2021survey}. The sub-domain under dialogue summarization domain, Meeting-based summarization, will be explained further in section \ref{sec:meetingsummarizers}. 

\textbf{Chat Summarization}: There is a growing interest in chat summarization because of the extensive usage of chat applications like Slack\footnote{https://slack.com/}, Discord\footnote{https://discord.com/}, etc. One of the earliest work is a personalized summarization based on user profiles from group chats, like Slack channels by a chat assistant service, Collabot~\cite{tepper2018collabot}. Another early work, SAMSum~\cite{gliwa2019samsum}, is a benchmark dataset for abstractive dialogue summarization with 16k dialogues which has facilitated much research in this domain. \citeauthor{chen2020multi} extracted different views from the conversations using SAMsum and generated multi-view dialogue summary utilizing a conversation encoder and a multi-view decoder. The authors proposed a multi-view decoder that is a transformer-based decoder to integrate encoded representations from different views and further generate summaries. Most of the recent research focuses on improving the quality of abstractive summaries.
\citeauthor{lee2021capturing} proposes self-supervised strategies that focus on correcting the speaker names by a post correction model thereby generating a revised summary. To reduce hallucinations and to improve factual consistency \citeauthor{liu2021controllable, narayan2021planning, wu2021controllable} proposed conditioning the summary with personal named entities, entity chain, and summary sketch, respectively. Graph based dialogue interaction models are also studied in some works. Summarization based on topic words using a guided graph model is one such work~\cite{zhao2020improving}. Another work under this category is by \citeauthor{feng2020incorporating} who integrated common sense knowledge in a heterogeneous graph along with utterance and speaker nodes. To overcome the absence of sufficient data, \citeauthor{gunasekara2021summary}, using pretrained language models, worked on generating conversations  for a given summary by three approaches --- supervised, reinforced, and turn-by-turn utterance generation in a controlled manner. This augmented conversation dataset can be used to improve the performance of summarization models.
MediaSUM~\cite{zhu2021mediasum} is another dataset for abstractive dialogue summarization which is worth mentioning. This dataset has more than 450K interview transcripts procured from national Public radio (NPR) and CNN annotated with description of topics and interview overviews as summaries.
\\
\textbf{Email Summarization}: Initial works on email thread summarization focused on multiple approaches: extracting topic phrases~\cite{muresan2001combining}, forming message clusters based on groups of topics~\cite{newman2003summarizing}, extracting important sentences based on specific email-related features~\cite{rambow2004summarizing}, etc. Recently, some task-specific approaches have been proposed, such as automatic email subject-line generation~\cite{zhang2019email} and email action item extraction/To-Do item generation~\cite{mukherjee-etal-2020-smart}.
\\
\textbf{Medical conversation Summarization}: Medical conversation summarization is an important application of dialogue summarization. According to \citeauthor{feng2021survey}, summarization methods in medical domain prefer to use a combination of extractive and abstractive methods to generate an accurate summary.
\\
\textbf{Customer conversation Summarization}: In addition to these domains, dialogue summarization is popular in summarizing customer interactions. Some of the customer service summarization models use topic modeling to generate accurate summaries~\cite{liu2019automatic,zou2020topic}. One of the datasets under this category is TODSum~\cite{zhao2021todsum} which has a state knowledge of dialogues like user intents, preferences, etc. along with the dialogues. 

\subsection{Metrics for summary evaluation}
The most common standard evaluation metric used for text summarization is Rouge (Recall-Oriented Understudy for Gisting Evaluation) which is an automatic approach for measuring the quality of generated summaries. This metric compares a generated summary with a set of human-generated summaries by counting the number of overlapping ngrams between the generated summary and reference summaries~\cite{lin2004rouge}. F1 scores for R1(Rouge-1, unigram overlap), R2(Rouge-2, bigram overlap), and Rouge-L (longest common sequence) are some of the common reported metrics in most papers. BERTScore~\cite{zhang2019bertscore} is another automatic metric which calculates semantic similarity with the help of contextual embeddings from pretrained BERT.

\section{Meeting Summarizers}
\label{sec:meetingsummarizers}
In this section we describe various summarization approaches for generating a concise summary from the meetings transcripts. This section is divided into extractive and abstractive based summarization approaches.

\subsection{Extractive Meeting Summarization}

\begin{figure}[ht!]
  \centering
 \includegraphics[width=0.5\textwidth]{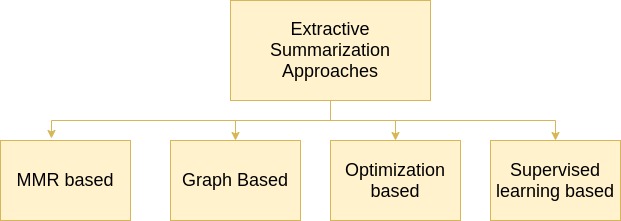}
 \caption{Extractive Summarization Approaches}
 \label{fig:diag1}
\end{figure}

One of the earliest work in extractive meeting summarization is by \citeauthor{murray2005extractive} that focused on a study of text summarization techniques like MMR and Latent Semantic Analysis (LSA) on ICSI dataset~\cite{janin2003icsi}. They have compared these approaches with feature-based summarization approaches that include prosodic and lexical features. The results showed that MMR and LSA outperformed the feature-based approaches. 
\citeauthor{riedhammer2008keyphrase} suggested using key phrases as queries to MMR can significantly improve the model performance. In addition to the features like lexical, structural, discourse, and prosodic features, \citeauthor{xie2008evaluating} found topic-related features as useful. In~\cite{xie2008using}, the authors proposed different ways to calculate similarity measures like centroid score and corpus-based similarity metrics~\cite{mihalcea2006corpus}. Numerous works were proposed afterwards that performed better than the greedy search algorithm, MMR. We can divide extractive summarizers into MMR-based, graph-based, optimization-based, and supervised categories~\cite{bokaei2016extractive} as shown in Figure \ref{fig:diag1}. 

The second category is graph-based algorithms where nodes represent utterances, and the edges represent the similarity between the utterances. A random walk on the graph is then used for ranking the utterances to be included in the generated summary. An unsupervised graph-based approach, ClusterRank~\cite{garg2009clusterrank} extended TextRank algorithm~\cite{mihalcea2004textrank} and performed better than MMR, taking care of redundant and off-topic data. In this approach, nodes are clusters of adjacent utterances grouped based on similarity, and the scoring of each utterance is based on the cluster it is associated with. The research work~\cite{bokaei2016extractive}, similar to ClusterRank, segments the meeting transcript using an unsupervised genetic algorithm and scores the utterances using a weighted scheme based on their association with each segment. The categorization of the segments is based on the participation of speakers which can be a monologue or discussion. A two layer graph consisting of utterance layer and speaker layer suggested by \citeauthor{chen2012two} uses three relations to score nodes namely, utterance-utterance, utterance-speaker, and speaker-speaker thereby giving importance to both speakers and utterances.

To overcome the non-optimality issue of MMR, optimization based approaches were introduced. Optimum utterances were selected using ILP algorithm~\cite{lin2009graph,gillick2009global,riedhammer2010long} under the constraint of summary length. The fourth category is based on supervised algorithms which performs a binary classification to decide whether to include an utterance in the summary or not. \citeauthor{xie2010improving} and \citeauthor{liu2010exploring} gave suggestions for improving supervised learning by incorporating different sampling techniques and speaker-related features, respectively. Prediction of summary worthy extracted dialogue acts (EDA) from the meeting corpora is another area which is explored by
\citeauthor{lai2013detecting}. They used a regression model using \emph{turn taking} features like \emph{participation equality}, \emph{turn taking freedom}, \emph{barge-in rate}, etc. for capturing various aspects of participation.

For predicting summary-worthy extracts, multimodal features like gestures, facial expressions, language, prosody, etc. are also important. \citeauthor{nihei2018fusing} discusses about fusing non-linguistic with linguistic features to perform a multimodal extractive summarization.
\citeauthor{murray2010generating} suggested that users prefer abstractive summaries over extractive summaries. They point that extractive summary, being a copy of the input text, copies the noise or grammatical mistakes present in the input document. Another weakness of extracive summaries is that they may loose their coherence. Thus, the user may need to review the original document to understand the context. Abstractive summaries are more preferable and are closer to human-generated summaries.

\subsection{Abstractive Meeting Summarization}
\begin{figure}[ht!]
  \centering
 \includegraphics[width=0.5\textwidth]{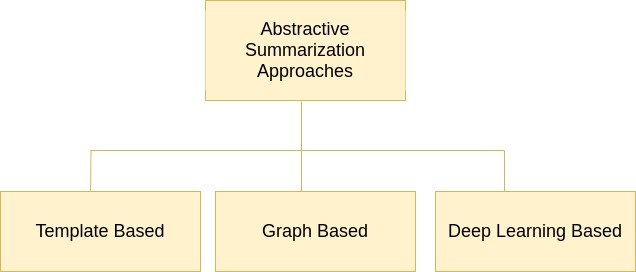}
 \caption{Abstractive Summarization Approaches}
 \label{fig:diag2}
\end{figure}

We can categorize abstractive summarization approaches as shown in Fig. \ref{fig:diag2}. 

\subsubsection{Graph Based Summarization}
 \citeauthor{mehdad2013abstractive} proposed detecting utterance's communities and applying entailment graph to select significant utterances. The final abstractive summary is obtained by applying word graph fusion algorithm based on ranking on these salient sentences.
\citeauthor{banerjee2015abstractive} used ILP based fusion approach to aggregate all the utterances that are important in each topic segment and generate a sentence summary for each topic segment. Utterances were represented as a dependency graph, and to increase the chance of merging, the author used anaphora resolution \footnote{Anaphora resolution is used to identify what a pronoun or noun phrase is referring to}. Anaphora resolution is used to replace pronouns with the original nouns from the previous utterance they referred to. This work was later extended in~\cite{banerjee2015generating}. The authors created an end-to-end framework starting from topic segmentation, identifying important utterances in each topic segment and then applying utterance fusion to obtain the summary. For topic segmentation, LCSeg~\cite{galley2003discourse} and bayesian unsupervised topic segmentation~\cite{eisenstein2008bayesian} methods were used, followed by an extractive summarization method to select important utterances. Contrary to their previous work, linguistic constraints were also introduced to the ILP problem to generate a readable summary. An unsupervised method using graph-based approach for abstractive summarization was proposed by \citeauthor{shang2018unsupervised}. In this approach, after performing the pre-processing step, the utterances are grouped together to form communities. These communities are based on a topic or a subtopic discussed in the meeting. Next, from each community, a single abstractive sentence is generated using an extension of Multi-sentence compression Graph~\cite{filippova2010multi}. The final step, is to select most important sentences from this category using budgeted submodular maximization.

\subsubsection{Template Based Summarization}
\begin{figure}[ht!]
  \centering
 \includegraphics[width=0.5\textwidth]{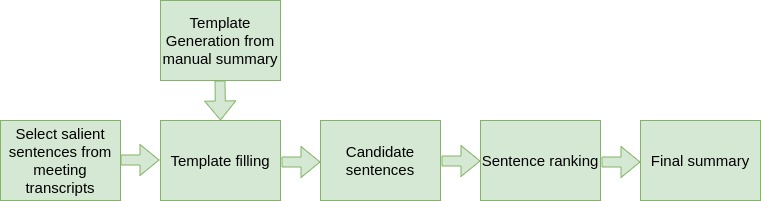}
 \caption{Workflow of template based summarization}
 \label{fig:template}
\end{figure}

The overall workflow of a template based summarization approach is shown in Figure \ref{fig:template}. The basic idea is to generate a template from the manual summary and fill these templates with significant information extracted from the input text. From these filled templates, sentences are selected to generate the final summary based on a ranking scheme. \citeauthor{wang2013domain} proposed focused abstractive meeting summary generation that summarizes a particular aspect of the meeting using templates. The first step on their framework was to identify dialogue act clusters. Salient information was selected from this cluster in the form of relation instances using a discriminative classifier based on support vector machines (SVM). The templates were learnt from manual summary using an algorithm based on Multiple Sequence Alignments (MSA)~\cite{durbin1998biological}. \emph{Overgenerate-and-Rank} strategy~\cite{walker2001spot,heilman2010good} was used for template filling followed by statistical ranking of a candidate abstract. The final summary was generated after removing redundancies. The paper~\cite{oya2014template} proposed template generation based on hypernym labeling, clustering, and applying a word graph-based fusion algorithm. Instead of a classifier, they used topic segmentation followed by extraction of important phrases and speakers. This work also used a novel template selection algorithm that identifies the templates associated with communities similar to the topic segments. 
Similarly, \citeauthor{singla2017automatic} also used template based abstractive summarization and explored the creation of communities using different heuristics.

\subsubsection{Deep Learning Based Summarization}
\begin{figure}[ht!]
  \centering
 \includegraphics[width=0.5\textwidth]{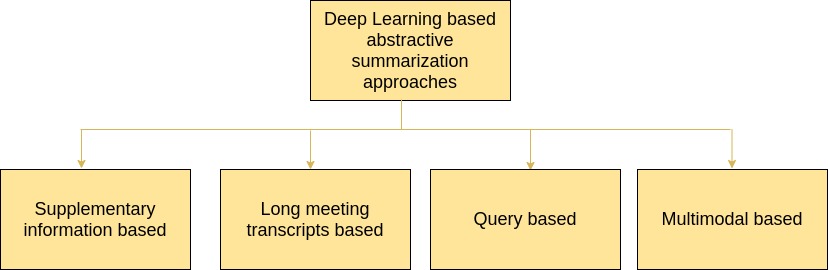}
 \caption{Deep learning based abstractive summarization approaches}
 \label{fig:diag3}
\end{figure}

We can see a paradigm shift in the abstractive summarization approaches after the evolution of neural networks. A categorization of deep learning-based summarization approaches is shown in Fig. \ref{fig:diag3}.

\paragraph{Query Based: }
According to~\cite{zhong2021qmsum}, meeting participants would like to get the summary of the meetings according to their area of interest. To serve this purpose the authors have proposed a summarization task based on a query. For this task they have also created a multi-domain summarization dataset with query-summary pairs. The summarization was done in a pipeline of two steps, locating the relevant span of meeting transcripts according to the query, and summarizing it with the help of state-of-art abstractive models like Pointer-Generator Network~\cite{see2017get}, BART~\cite{lewis-etal-2020-bart}, HMNet~\cite{zhu2020hierarchical}. The authors used two methods for locator, a Pointer Network~\cite{vinyals2015pointer} and a hierarchical ranking based model. \citeauthor{vig2021exploring} conducted an exploratory study of various approaches on Query Focused Summarization. The authors also proposed two end-to-end neural network-based approaches which are composed of extractor models followed by abstractor models.

\paragraph{Multi-modal based:}
The lack of structure alongside grammatical mistakes of meeting data makes the natural language models to lose focus in summarizing. \citeauthor{li2019keep} proposed using multi-modal data in addition to the meeting transcripts for the task of summarization. They proposed to use Visual Focus Of Attention (VFOA) as a feature to select salient data. The work also proposes a multi-modal hierarchical attention which focuses on topic segments, and then to the utterances and words in it. VFOA is obtained by observing the participants' video recordings for each frame and estimate the eye gaze direction vector and head pose angle. If a speaker receives more attention, we can consider their utterances as a summary-worthy utterance which should be paid more attention when generating a summary. Topic segmentation and summarization model are trained jointly in this work.
\paragraph{Summarization Models for Long meetings:}
There are a set of works which focus on summarization of long meeting transcripts. One of the methods enabling sequence-to-sequence models to summarize long transcripts is to use hierarchical attention. \citeauthor{zhao2019abstractive} recommended a hierarchical encoder-decoder network with adaptive segmentation of conversations for the summarization task. The semantic representation of meeting transcripts is learned by the encoder which is a bi-directional LSTM along with the conversation segmentation.
A decoder trained with reinforcement learning framework is used to generate the abstractive summary. In~\cite{zhu2020hierarchical}, the authors have discussed about using a hierarchical structure for handling long meeting transcripts and proposing a Hierarchical Meeting summarization Network (HMNet) which is based on transformer architecture~\cite{vaswani2017attention}. The attention is given in the token level and turn level of the meeting dialogues, and the role vector of speakers are added to the turn level encoding. To overcome the challenge of inadequate meeting summary data, this model is pretrained on various news summarization datasets. Here, the data from news domain is converted into the meeting format where different news articles mimic a multi-person meeting. Each sentence forms a turn and they are reshuffled to resemble a mixed order of speakers. \citeauthor{rohde2021hierarchical} designed a Hierarchical Attention transformer (HAT) architecture by modifying the standard transformer architecture~\cite{vaswani2017attention} and adding hierarchical attention. They added BOS tokens at the starting of each sentence and added a layer in the encoder block which attends to these BOS embeddings. This layer generates contextual representations for the BOS tokens which can be considered as sentence level representations. An additional attention module is added to the decoder to attend these BOS token embeddings. \citeauthor{zhang2021exploratory} performed a comparative study of three different methods for long dialogue summarization. The three models are (1) Longformer --- the extended transformer model~\cite{beltagy2020longformer} ---, (2) \emph{retrieve-then-summarize} pipeline models, and (3) the hierarchical HMNet model. Their study established that \emph{retrieve-then-summarize} pipeline performs well compared to other models. Two main stages of the pipeline are retrieving relevant utterances and summarization. For retrieval, they experimented with TF-IDF~\cite{jones1972statistical}, BM25~\cite{robertson2009probabilistic}, and Locator~\cite{zhong2021qmsum}, and for summarization, BART-large model~\cite{lewis-etal-2020-bart} was used. Some of the works focused on sliding window approach in which the long transcripts are broken down into small windows and a summary is generated from each window. \citeauthor{koay2021sliding} proposed a sliding window approach along with a Bart based abstractive summarizer to obtain the partial summaries of each window. Using these local summaries, salient utterances were identified based on Rouge score. These selected utterances were then included to the main meeting summary. The results were promising when a sliding window approach was used. \citeauthor{liu2021dynamic} also proposed a dynamic sliding window approach where the window stride size were predicted by their proposed model called Retrospective. The decoder predicts the context boundary along with the summary sentence. The final summary is obtained by concatenating the summary of each window segment. \citeauthor{zhang2021summ} suggested a new framework for handling long input data called Summ\^{} N which had a coarse-grained and fine-grained stage. The coarse-grained stage performed the compression of the input text by doing a segmentation followed by a coarse summary generation. Multiple coarse stages were repeated with different model parameters. Finally, the coarse summary is passed to the fine-grained stage to get the final summary. The underlying summarizer model used is BART~\cite{lewis-etal-2020-bart}. In another study, a pre-trained model called DIALOGLM was suggested in~\cite{zhong2021dialoglm} for understanding and summarizing long dialogues. The proposed model is a neural auto-encoder, and a window-based denoising method is used for pretraining the model. For this purpose, five types of noise are introduced based on the dialogue characteristics. The model is trained to restore the original content out of the given noisy content and the remaining dialogue. After pretraining, the model is applied on three downstream tasks including long dialogue summarization.

\paragraph{Supplementary Information based: }
Some of the works have used supplementary information as a guidance for creating summarizer models. RetrievalSum~\cite{an2021retrievalsum} retrieves high quality semantically-similar exemplars from a knowledge base which acts as a guidance of writing format and also provides background information. Given the query document, the retrieval module retrieves exemplars by a semantic matching process. \citeauthor{ganesh2019restructuring} and \citeauthor{feng2020dialogue} incorporated discourse relations into their summarizer pipeline. \citeauthor{ganesh2019restructuring} proposed a zero-shot learning based abstractive summarization method using discourse relations. These discourse relations are used to restructure the conversations into a document. This is followed by a document summarization process using two models: pointer-generator networks~\cite{see2017get} and transformer-based model BART~\cite{lewis-etal-2020-bart}. Discourse labels generated using Conditional Random fields~\cite{okazaki2007crfsuite} is used for dialogue restructuring into a document. \citeauthor{feng2020dialogue} presented a Dialogue Discourse-Aware Meeting Summarizer (DDAMS) that models the dialogue discourse relations among utterances. The meeting utterances with discourse relations are converted into a meeting graph which is then modeled using a graph encoder. \citeauthor{koay2020domain} annotated the dataset with domain terminologies (jargon terms) and studied their impact on summarizers performance. \citeauthor{goo2018abstractive} leveraged dialogue acts for dialogue summarization using sentence-gated modeling. Dialogue act is based on the interactive patterns between speakers which can be very informative for sumarization.
In a different study, \citeauthor{beckage2021context} explored incremental temporal summarization with a focus on previous summaries generated by humans. They also suggest semantic role labelling to extract information from past summaries. 

\begin{table*}[]

\centering
\resizebox{\linewidth}{!}{%
\begin{tabular}{ccccccc}
\hline
\multicolumn{1}{c|}{\textbf{}} & \multicolumn{3}{c|}{\textbf{AMI}} & \multicolumn{3}{c}{\textbf{ICSI}} \\
\multicolumn{1}{c|}{\textbf{Model}} & \textbf{R-1} & \textbf{R-2} & \multicolumn{1}{c|}{\textbf{R-L}} & \textbf{R-1} & \textbf{R-2} & \textbf{R-L} \\ \hline
\multicolumn{7}{c}{\textbf{Extractive Methods}} \\ \hline
\multicolumn{1}{c|}{TextRank (\citeauthor{mihalcea2004textrank})} & 35.19 & 6.13 & \multicolumn{1}{c|}{15.7} & 30.72 & 4.69 & 12.97 \\
\multicolumn{1}{c|}{SummaRunner (\citeauthor{nallapati2017summarunner})} & 30.98 & 5.54 & \multicolumn{1}{c|}{13.91} & 27.6 & 3.7 & 12.52 \\ \hline
\multicolumn{7}{c}{\textbf{Abstractive Methods (DL-based)}} \\ \hline
\multicolumn{1}{c|}{Topicseg (\citeauthor{li-etal-2019-keep}} & 51.53 & 12.23 & \multicolumn{1}{c|}{25.47} & - & - & - \\
\multicolumn{1}{c|}{Topicseg+vfoa (\citeauthor{li-etal-2019-keep})} & 53.29 & 13.51 & \multicolumn{1}{c|}{26.9} & - & - & - \\
\multicolumn{1}{c|}{HMNet(\citeauthor{zhu2020hierarchical})} & 53.02 & 18.57 & \multicolumn{1}{c|}{-} & 46.28 & 10.6 & - \\
\multicolumn{1}{c|}{HAT-CNNDM(\citeauthor{rohde2021hierarchical})} & 52.27 & 20.15 & \multicolumn{1}{c|}{50.57} & 43.98 & 10.83 & 41.36 \\
\multicolumn{1}{c|}{Sliding Window(\citeauthor{koay2021sliding})} & 51.2 & 27.6 & \multicolumn{1}{c|}{-} & - & - & - \\
\multicolumn{1}{c|}{BART-SW-Dynamic(\citeauthor{liu2021dynamic})} & 52.83 & 21.77 & \multicolumn{1}{c|}{26.01} & - & - & - \\
\multicolumn{1}{c|}{SUMM\textasciicircum{}N(\citeauthor{zhang2021summ})} & 53.44 & 20.3 & \multicolumn{1}{c|}{51.39} & 48.87 & 12.17 & 46.38 \\
\multicolumn{1}{c|}{DIALOGLM (l = 5, 120)(\citeauthor{zhong2021dialoglm})} & 54.49 & 20.03 & \multicolumn{1}{c|}{51.92} & 49.25 & 12.31 & 46.8 \\
\multicolumn{1}{c|}{DIALOGLM-sparse (l = 8192)(\citeauthor{zhong2021dialoglm})} & 53.72 & 19.61 & \multicolumn{1}{c|}{51.83} & 49.56 & 12.53 & 47.08 \\
\multicolumn{1}{c|}{Sentence-Gated (\citeauthor{goo2018abstractive})} & 49.29 & 19.31 & \multicolumn{1}{c|}{24.82} & 39.37 & 9.57 & 17.17 \\
\multicolumn{1}{c|}{Retrievalsum (\citeauthor{an2021retrievalsum})} & 56.26 & 34.9 & \multicolumn{1}{c|}{52.51} & - & - & - \\
\multicolumn{1}{c|}{DDAMS(\citeauthor{feng2020dialogue})} & 53.15 & 22.32 & \multicolumn{1}{c|}{25.67*} & 40.41 & 11.02 & 19.18 \\
\multicolumn{1}{c|}{Domain Terminology(\citeauthor{koay2020domain})} & - & - & \multicolumn{1}{c|}{-} & 60.7 & 37.1 & - \\
\multicolumn{1}{c|}{Longformer-BART (\citeauthor{fabbri2021convosumm})} & 54.2 & 20.72 & \multicolumn{1}{c|}{51.36} & 43.03 & 12.14 & 40.26 \\
\multicolumn{1}{c|}{Longformer-BART-arg (\citeauthor{fabbri2021convosumm})} & 54.47 & 20.83 & \multicolumn{1}{c|}{51.74} & 44.17 & 11.69 & 41.33 \\ \hline
\end{tabular}%
}
\caption{Leaderboard. We adopt reported results from literature. Results with $*$ indicate that ROUGE-L is calculated without sentence splitting. }
\label{tab:table-1}
\end{table*}

\section{Datasets }
\label{sec:datasets}
AMI~\cite{carletta2005ami} and ICSI~\cite{janin2003icsi} are the two main corpora used for meeting summarization task. The meeting transcripts are obtained using Automatic Speech Recognition (ASR). Both of the datasets contain human annotated extractive and abstractive summaries. The AMI dataset meetings are about the product design of a new remote controller. The ICSI dataset however contains academic discussions between research students. AMI and ICSI datasets have 289 and 464 turns with 4757 and 10,189 words, respectively~\cite{zhu2020hierarchical}. Each summary in AMI and ICSI datasets contains 322 and 534 words on average, respectively. QMSum~\cite{zhong2021qmsum} is a multi-domain dataset which contains data from three domains such as academic meetings, product meetings, and committee meetings. It has 1,808 query-summary pairs over 232 meetings --- 137 AMI meetings, 59 ICSI meetings, and 36 Parliament committee meetings. ConvoSumm~\cite{fabbri2021convosumm} is a suite of four datasets of online conversations based on news comments, discussion forums, community question answering forums, and email threads. Using these datasets, the authors also have benchmarked state-of-the-art summarization models. In this survey, we try to cover all publicly reported results for any of these datasets.

\section{Discussions}
\label{sec:Discussions}
The leaderboard of the meeting summarizers is shown in the table ~\ref{tab:table-1}. The results are reported from published literature reviews including \cite{feng2021survey}, \cite{fabbri2021convosumm}, and \cite{zhong2021dialoglm}, and the original papers. Missing values in the leaderboard indicate that the results are not available for the corresponding dataset. Besides, in all of these cases, the code is not publicly made available to reproduce the experiments on the other datasets for which results are not reported. For dataset AMI, the Rouge-1 score of RetrievalSum is the highest($56.26$) when compared to other deep learning based abstractive methods. Models such as DIALOGLM and Longformer-BART-arg follows RetrievalSum in the Rouge-1 score with the values $54.49$ and $54.47$. The highest Rouge-1 score for RetrievalSum can be attributed to the two mechanisms --- Group Alignment and Rouge Credit which helps to learn the structure and style of summary from exemplars in addition to the background knowledge. DIALOGLM cannot be compared with RetrievalSum as the former is a pretrained model specifically designed for long dialogue understanding and summarization. For ICSI dataset, DIALOGLM outperforms all other models with a ROUGE-1 score of $49.56$. For example, this model exceeds ROUGE-1 score of HMNet by $2.97$ points. Longformer-BART-arg is a longformer model initialised with BART parameters and trained with argument-mining based input. This model allows up to 16K tokens in the text input. The results of this model are comparable to those of DIALOGLM for AMI but the latter performs better for ICSI dataset. For ICSI dataset, \cite{koay-etal-2020-domain} outperforms every other model with ROUGE-1 value of $60.7$, despite its simplicity. The results show the impact of domain terminology on meeting summarization.

\section{Challenges}
\label{sec:challenges}
An abstractive text summarizer often suffers from a factual inconsistency problem. These problems are also called as hallucinations. Meeting summarizers also can suffer from this issue. Besides, lengthy meeting transcripts often pose a challenge to the meeting summarizers. The number of tokens in a meeting transcript is typically more than what can be handled by a transformer architecture. Many recent works have focused on this issue trying to effectively summarize lengthy inputs. However, this still is a challenge which needs more attention. In addition to this, there is not enough annotated datasets for meetings as most of the meetings performed in industry are proprietary in nature. Lack of structure is a problem faced in meeting summarization task. There are multiple participants in a meeting and the information is distributed across dialogues between the speakers. The dialogues are more verbose in nature and also suffer from lack of coherency.
 
\section{Conclusion }
\label{sec:conclusion}
The pandemic era resulted in a rise of remote working and the usage of video conferencing tools. These virtual meetings could be transcribed easily with the help of ASR systems, thereby increasing the demand for automatic summarization systems for meetings. This paper thoroughly surveys meeting summarization work that have happened till date. This literature covers the state-of-the-art extractive summarization and abstractive summarization models for meetings. We also discuss the challenges faced by researchers and shed light into future research work. Some of the future works can focus on the factual inconsistency problem and the summarization of lengthy meeting transcripts. Exploring the the impact of speakers' roles in the generated summary could also be a potential future direction for researchers.

\bibliography{anthology,custom}
\bibliographystyle{acl_natbib}



\end{document}